\documentclass[12pt,a4paper]{article}

\usepackage[american]{babel}

\usepackage[a4paper,top=2cm,bottom=2cm,left=2.5cm,right=2.5cm,marginparwidth=1.75cm]{geometry}

\usepackage[numbers]{natbib}

\newcommand{\cellWithForcedBreak}[2][c]{
    \begin{tabular}[#1]{@{}c@{}}#2\end{tabular}
}

\newcommand{\cellWithForcedBreakLeft}[2][c]{
    \begin{tabular}[#1]{@{}l@{}}#2\end{tabular}
}

\usepackage{graphicx}
\usepackage{multirow}
\usepackage{multicol}
\usepackage{amsmath}
\usepackage{graphicx}
\usepackage[colorlinks=true, allcolors=blue]{hyperref}
\usepackage{hyperref}
\usepackage[title]{appendix}
\usepackage{mathrsfs}
\usepackage{amsfonts}
\usepackage{booktabs} 
\usepackage{caption}  
\usepackage{threeparttable} 
\usepackage{algorithm}
\usepackage{algorithmicx}
\usepackage{algpseudocode}
\usepackage{listings}
\usepackage{enumitem}
\usepackage{chngcntr}
\usepackage{booktabs}
\usepackage{lipsum}
\usepackage{subcaption}
\usepackage{authblk}
\usepackage[T1]{fontenc}    
\usepackage{csquotes}       
\usepackage{diagbox}
\usepackage{comment}
\usepackage{glossaries-extra} 

\usepackage{setspace}
\usepackage{titlesec}
\titleformat{\section} 
  {\normalfont\Large\bfseries}{\thesection.}{1em}{}
  
\usepackage{lineno} 

\rightlinenumbers 

\usepackage{float}   
\usepackage{caption} 


\date{}

\title{Medicine on the Edge: Comparative Performance Analysis of On-Device LLMs for Clinical Reasoning}

\author[1,2,*]{Leon Nissen}
\author[1]{Philip Zagar}
\author[1]{Vishnu Ravi}
\author[1]{Aydin Zahedivash}
\author[2]{Lara Marie Reimer}
\author[2]{Stephan Jonas}
\author[1]{Oliver Aalami}
\author[1]{Paul Schmiedmayer}
\affil[1]{\small Stanford Mussallem Center for Biodesign, Stanford University, 318 Pasteur Drive, Stanford, California, USA}
\affil[2]{\small Institute for Digital Medicine, University Hospital Bonn, Venusberg-Campus 1, Germany}
\affil[*]{\small Corresponding author. \texttt{leon.nissen@stanford.edu}}

\makeglossaries
\newabbreviation{LLM}{LLM}{large language model}
\newabbreviation{IRB}{IRB}{Institutional Review Board}
\newabbreviation{TTFT}{TTFT}{time to first token}
\newabbreviation{TPS}{T/s}{tokens per second}
\newabbreviation{GB}{GB}{gigabytes}
\newabbreviation{CPU}{CPU}{central processing unit}
\newabbreviation{GPU}{GPU}{graphics processing unit}
\newabbreviation{SoC}{SoC}{system on a chip}
\newabbreviation{API}{API}{application programming interface}
\newabbreviation{FHIR}{FHIR}{Fast Healthcare Interoperability Resources}

\begin{document}
\maketitle

\begin{abstract}
The deployment of \glspl*{LLM} on mobile devices offers significant potential for medical applications, enhancing privacy, security, and cost-efficiency by eliminating reliance on cloud-based services and keeping sensitive health data local.
However, the performance and accuracy of on-device \glspl*{LLM} in real-world medical contexts remain underexplored.
In this study, we benchmark publicly available on-device \glspl*{LLM} using the AMEGA dataset, evaluating accuracy, computational efficiency, and thermal limitation across various mobile devices.
Our results indicate that compact general-purpose models like Phi-3 Mini achieve a strong balance between speed and accuracy, while medically fine-tuned models such as Med42 and Aloe attain the highest accuracy.
Notably, deploying \glspl*{LLM} on older devices remains feasible, with memory constraints posing a greater challenge than raw processing power.
Our study underscores the potential of on-device \glspl*{LLM} for healthcare while emphasizing the need for more efficient inference and models tailored to real-world clinical reasoning.
\end{abstract}

\textbf{Keywords}: Digital Health, LLM, Benchmark, On-Device


\section{Introduction}
\label{sec:intro}
As the adoption of \glspl*{LLM} expands, the growing demand has driven the development of smaller, open-weight, and cost-effective models~\cite{fourrier_2023_2023}.
These models enable fine-tuning, quantization, and inference of \glspl*{LLM} on edge devices such as smartphones and tablets~\cite{dhar_empirical_2024}.
Deploying \glspl*{LLM} on edge devices enhances privacy~\cite{wang_privatelora_2023}, which is particularly critical for applications involving sensitive medical data~\cite{gostin_health_2018}.
Additionally, distributing model inference across numerous user devices reduces overall energy consumption~\cite{mehta_how_2024}.
On-device inference also eliminates cloud costs by leveraging users' existing hardware for model inference.

Despite these advantages, the real-world applicability of medical \glspl*{LLM} on edge devices remains underexplored.
While model repositories such as Hugging Face offer a diverse selection of fine-tuned \glspl*{LLM} for medical applications\footnote{\url{https://huggingface.co/models?other=medical&sort=created}}, these models are typically trained on multiple-choice medical exam datasets, which do not accurately reflect the complexities of real-world clinical decision-making~\cite{raji_its_2025, gu_probabilistic_2024}.
To address this limitation, robust benchmarking methods are needed to assess \glspl*{LLM}'s problem-solving capabilities in real-world diagnostic scenarios, including open-ended questions.
\citeauthor{fast-2024-amega-llm} introduced a benchmark with a dataset comprising 20 clinical cases across 13 medical specialties.
This dataset evaluates \glspl*{LLM}'s ability to follow medical guidelines in realistic clinical scenarios, emphasizing diagnostic reasoning, treatment planning, and adherence to established protocols.
It aims to replicate real-world doctor-patient interactions using open-ended question-answering formats rather than traditional multiple-choice questions.

This paper evaluates mobile, on-device \glspl*{LLM} for clinical case reasoning using the AMEGA medical question benchmark.
We assess accuracy, compatibility, performance, and feasibility across mobile devices with varying computational capabilities.
Our analysis benchmarks state-of-the-art \glspl*{LLM} and edge device capabilities for medical applications, highlighting accuracy-efficiency trade-offs.

\section{Methods}
\label{sec:methods}

To evaluate the performance and real-world usability of \glspl*{LLM} on resource-constrained smartphones in a medical context, we developed the \textit{HealthBench} iOS application, which supports both iPhone and iPad devices within the Apple ecosystem.
Using the open-source AMEGA benchmarking dataset, we assessed on-device \gls*{LLM} inference across devices with varying resource capabilities.
We tested both emerging and widely adopted \glspl*{LLM}, with and without a medical focus, to comprehensively compare lower-quantized model variants suitable for on-device inference~\cite{xiao2024smoothquantaccurateefficientposttraining}.

The HealthBench application employs the open-source Stanford \textit{Spezi LLM} module\footnote{\url{https://github.com/StanfordSpezi/SpeziLLM/}}, which leverages Apple's \textit{MLX} framework~\cite{mlx2023} for \gls*{LLM} inference.
MLX is an array framework for machine learning, specifically optimized for Apple Silicon \gls*{SoC}\footnote{\url{https://developer.apple.com/documentation/apple-silicon}}.
SpeziLLM utilizes \textit{MLX-Swift}\footnote{\url{https://github.com/ml-explore/mlx-swift}}, a Swift-based extension of MLX, along with \textit{MLX-Swift-Examples}\footnote{\url{https://github.com/ml-explore/mlx-swift-examples}}, as abstraction layers to orchestrate \gls*{LLM} inference.

\subsection{Dataset}
\label{sec:dataset}
The AMEGA benchmark dataset, published in 2024, consists of 20 carefully selected clinical cases featuring open-ended questions designed to evaluate \glspl*{LLM}' adherence to medical guidelines~\cite{fast-2024-amega-llm}.
We chose this benchmark for its open-ended question format and fully automatable, reproducible evaluation strategy, reducing human bias.  
Additionally, the AMEGA question style enhances real-world generalizability by avoiding the benchmarking of \glspl*{LLM} with multiple-choice medical exam questions~\cite{raji_its_2025}.  
AMEGA requires models to generate text-based responses rather than assigning probabilities to fixed candidate labels found in medical exams.

Each case presents a detailed patient vignette with six to eight associated questions, covering 20 medical specialties, including oncology, cardiology, and emergency medicine.  
All 20 cases are scored on a scale from 0 (worst) to 50 (best), resulting in an overall score between 0 (worst) and 1000 (best) for each model-device combination. 

In addition to individual questions, the benchmark incorporates a \textit{reask} process~\cite{fast-2024-amega-llm}, allowing models to refine their initial responses if they do not fully meet the criteria.
However, the reask process significantly increases response times and can lead to suboptimal results with smaller models, which are more prone to hallucinations or non-meaningful answers~\cite{krishna_importing_2025}.
These insights were observed during our pilot phase but require further validation.
Our preliminary findings suggest that including the reask process for on-device models extends evaluation time without providing meaningful improvements over a simple zero-shot prompting approach.
However, we aim to further investigate this phenomenon and the reask process on-device in future work.
Consequently, we excluded the reask process from our benchmark while still enabling direct comparison with initial results from larger models leveraging greater computational resources.

\subsection{Metrics}
\label{sec:metrics}
During benchmarking, where all cases and their respective questions are processed sequentially, the HealthBench app monitors various device performance metrics, including \gls*{CPU} usage, memory consumption, thermal state, and battery level.  
Performance data is logged every 500 milliseconds.
The application records model-specific metrics for each question, such as \gls*{TTFT} and \gls*{TPS}.

Battery levels are monitored to analyze energy consumption resulting from model inference.  
Since the release of iOS 17, battery level reporting on iPhones and iPads has been limited to a resolution of 5\%\footnote{\url{https://forums.developer.apple.com/forums/thread/732903}}, reducing monitoring accuracy but potentially improving privacy, as battery level data can be used for tracking~\cite{hern_your_2016}.  

The thermal state\footnote{\url{https://developer.apple.com/documentation/foundation/processinfo/thermalstate}}, an aggregate measure based on multiple temperature sensor readings, is classified into four levels: \textit{nominal} (within normal limits), \textit{fair} (slightly elevated), \textit{serious} (high), and \textit{critical} (significantly impacting performance and requiring a device cool-down).
A cool-down phase is initiated before inference for each questin begins if the thermal state reaches \textit{critical} to mitigate performance degradation.
However, no cool-down is initiated during inference, regardless of the thermal state.

After each question, the application clears the GPU cache, which "\textit{causes all cached metal buffers to be deallocated}"\footnote{\url{https://swiftpackageindex.com/ml-explore/mlx-swift/main/documentation/mlx/gpu/clearcache()}}, preventing memory pressure and potential application termination.
This ensures that individual question runs remain independent, minimizing bias between them.
This step is crucial, as the GPU cache, managed by the MLX framework, can grow by several gigabytes during benchmarking if not cleared.

A low-temperature model setting was selected to enhance reproducibility.  
However, temperatures below 0.1 were avoided, as preliminary testing indicated that smaller models frequently produce illogical outputs at such levels.
The output token length was limited to 2,048 for standard models and 4,096 for reasoning models to mitigate repetitive outputs and maintain a manageable memory footprint.

The application was deployed as an optimized \texttt{xcodebuild} release build without an attached \texttt{lldb} debugger to ensure performance accurately reflects real-world usage conditions.

\subsection{Model Selection}
\label{sec:model-selection}

A key requirement was the model's compatibility with the MLX format to enable on-device inference via Apple's MLX framework.  
The \gls*{LLM} selection is a convenience sample based criteria like model parameter size, static memory usage, type (medical vs. non-medical), reasoning vs. non-reasoning models, and the goal of covering the most suitable and up-to-date models.
The selected open-weight DeepSeek reasoning models are distilled models based on both the Qwen2 and Llama base models.

The static memory usage of a large language model refers to the memory required to load its parameters and embeddings into memory before any inference occurs, excluding context-related allocations such as the key-value (KV) cache.
This state represents the model's idle memory usage, providing a baseline for hardware requirements independent of runtime context expansion.  
Static memory usage was particularly crucial, as even the iPhone 15, released in September 2023, is equipped with only 6~\gls*{GB} of memory, limiting its capacity to support larger models.

All models in the evaluation (\autoref{tab:model-selection}) were converted to the model format required by the MLX framework, either by the authors or the broad MLX community.
The \texttt{quantize} parameter was enabled during the conversion to ensure broad device compatibility.
The default MLX conversion settings were used, with a quantization group size of 64, 4 quantization bits, and a \texttt{float64} data type\footnote{\url{https://github.com/ml-explore/mlx-examples/blob/main/llms/llama/convert.py}}.

During initial testing, we identified an issue with the MedFound~8B model: its configuration file lacked a defined chat template, leading to degenerate \gls*{LLM} outputs.
To mitigate this, we experimented with various chat templates, including the Llama3 template, but the issue persisted.
Although applying the model’s recommended tokenizer provided partial improvement, preliminary testing indicated potential signs of memorization from the training data.
Due to these limitations, we excluded the MedFound~8B model from on-device benchmarking, resulting in a final evaluation set of 13 models.

Both general-purpose models and fine-tuned variants~\cite{jeong2024finetuning} were utilized for evaluation to ensure a comprehensive performance assessment. The \textit{Model Type} denotes whether a model was explicitly fine-tuned for medical applications. All specialized medical models in this study are fine-tuned versions of pre-trained foundation models.
Fine-tuning refers to the process of further training a general-purpose model on domain-specific data to adapt its parameters for specialized tasks.  
This approach enhances the model's domain knowledge and optimizes its performance for specific applications.  
Similar to the large-scale AMEGA benchmark, we aimed to determine how fine-tuned (medical) models compare to general-purpose (non-medical) models~\cite{fast-2024-amega-llm}.

Finally, we classified the models into three sizes which are defined by the number of parameters: \textit{small} ($[1, 3)$ billion), \textit{medium} ($[3, 7)$ billion), and \textit{large} ($[7, \infty)$ billion).

\begin{table*}[h]
    \centering
    \makebox[\textwidth]{
    \resizebox{1.15\textwidth}{!}{
    \begin{tabular}{ccccccc}
      \toprule
      \textbf{Name} & \bfseries Base Model & \cellWithForcedBreak{\textbf{Model} \\ \textbf{Parameters} \\ (Billion)} & \cellWithForcedBreak{\textbf{Model} \\ \textbf{Type}} & \cellWithForcedBreak{\textbf{Quantization} \\ (Original/MLX)} & \cellWithForcedBreak{\bfseries Context \\ \textbf{Window} \\ (Tokens)} & \cellWithForcedBreak{\textbf{Memory} \\ \textbf{Usage} \\ (GB)}\\
      \midrule
      \cellWithForcedBreak{BioMedical 1B\\\cite{ContactDoctor_Bio-Medical-Llama-3.2-1B-CoT-012025}} & llama3.2-instruct 1B & 1 (small) & Medical & 4/4 & 131,072 & 1.20 \\
      \cellWithForcedBreak{Llama 3.2 1B Instruct\\\cite{MetaLlama_Llama-3.2-1B-Instruct}} & - & 1 (small) & Non-Medical & 4/4 & 131,072 & 1.10 \\
      \cellWithForcedBreak{DeepSeek R1 1.5B\\\cite{deepseekai2025deepseekr1incentivizingreasoningcapability}} & qwen2.5-math 1.5B & 1.5 (small) & Non-Medical & -/8 & 131,072 & 2.24 \\
      \cellWithForcedBreak{BioMedical 3B\\\cite{ContactDoctor_Bio-Medical-3B-CoT-012025}} & qwen2.5-instruct 3B & 3 (medium) & Medical & -/4 & 32,768 & 2.10 \\
      \cellWithForcedBreak{Llama 3.2 3B Instruct \\\cite{MetaLlama_Llama-3.2-3B-Instruct}} & - & 3 (medium) & Non-Medical & 4/4 & 131,072 & 2.11 \\
      \cellWithForcedBreak{Phi 3 mini 3.8B Instruct\\\cite{Microsoft_Phi-3-mini-4k-instruct}} & - & 3.8 (medium) & Non-Medical & -/4 & 4,096 & 2.36 \\
      \cellWithForcedBreak{DeepSeek R1 7B\\\cite{deepseekai2025deepseekr1incentivizingreasoningcapability}} & qwen2.5-math 7B & 7 (large) & Non-Medical & -/4 & 131,072 & 4.42 \\
      \cellWithForcedBreak{Qwen 2 7B\\\cite{qwen2}} & - & 7 (large) & Non-Medical & -/4 & 32,768 & 4.36 \\
      \cellWithForcedBreak{BioMedical 8B\\\cite{ContactDoctor_Bio-Medical-Llama-3-8B}} & llama3.2-instruct 8B & 8 (large) & Medical & -/4 & 8,192 & 4.69 \\
      \cellWithForcedBreak{Aloe (Beta) 8B\\\cite{gururajan2024aloe}} & llama3.1-instruct 8B & 8 (large) & Medical & -/4 & 131,072 & 4.69 \\
      \cellWithForcedBreak{MedLlama 8B\\\cite{YonseiMAILAB_medllama3-v20}} & llama3.0-instruct 8B & 8 (large) & Medical & -/4 & 8,192 & 4.69 \\
      \cellWithForcedBreak{Med42 8B\\\cite{med42v2}} & llama3.0-instruct 8B & 8 (large) & Medical & -/4 & 8,192 & 4.69 \\
      \cellWithForcedBreak{Llama 3.1 8B Instruct\\\cite{MetaLlama_Llama-3.1-8B-Instruct}} & - & 8 (large) & Non-Medical & -/4 & 131,072 & 4.58 \\
      \cellWithForcedBreak{DeepSeek R1 8B\\\cite{deepseekai2025deepseekr1incentivizingreasoningcapability}} & llama-3.1 8B & 8 (large) & Non-Medical & -/4 & 131,072 & 4.69 \\
      \bottomrule
\end{tabular}}}
\caption{\textbf{Used LLMs for benchmarking.} The table outlines the name, base model, model size (in billions of parameters), specialization (Medical or Non-Medical fine-tuning), quantization of the original model and the converted MLX model (where "-" denotes that the model was not quantized), context window size (in tokens), and memory usage (in \gls*{GB}).}
\label{tab:device-selection}
\end{table*}

\subsection{Devices}
\label{sec:devices}

To represent real-world usage scenarios, we selected a diverse range of widely available mobile devices supported by the MLX framework, featuring varying resource capacities.
MLX utilizes the entire \gls*{SoC}, including the \gls*{CPU}, \gls*{GPU}, and unified memory, which is accessible by both the \gls*{CPU} and \gls*{GPU}.
Consequently, when running a \gls*{LLM} with MLX, the model operates on the \gls*{GPU} and fully leverages the chip architecture for optimal performance.

\begin{table*}[h]
    \centering
    \makebox[\textwidth]{
        \resizebox{1.15\textwidth}{!}{
        \begin{tabular}{ccccccc}
        \toprule
        \bfseries Name & \bfseries Identifier & \cellWithForcedBreak{\bfseries Release \\ \bfseries{Year}} & \cellWithForcedBreak{\bfseries SoC} & \cellWithForcedBreak{\bfseries Memory \\ (GiB)} & \cellWithForcedBreak{\textbf{Memory} \\ \textbf{Limit} (GiB)} & \cellWithForcedBreak{\bfseries Battery Max. \\ \textbf{Capacity} (\%)}\\
          \midrule
          iPhone 12 Pro & iPhone13,3 & 2020 & A14 Bionic & 6 & 3.94 & 79 \\
          iPhone 13 & iPhone14,5 &  2021 & A15 Bionic & 4 & 2.25 & 88\\
          iPhone 15 Pro & iPhone16,1 & 2023 & A17 Pro & 8 & 5.94 & 89\\
          iPhone 16 Pro Max & iPhone17,2 & 2024 & A18 Pro & 8 & 5.94 & 100\\
          iPad Pro (2022) & iPad14,3 & 2022 & M2 & 8 & 7.93 & 97\\
          iPad Pro (2024) & iPad16,3 & 2024 & M4 & 16 & 15.90 & 100\\
          \bottomrule
        \end{tabular}}
    }
    \caption{Device selection for the benchmark, including the device model characteristics and battery health.}
    \label{tab:model-selection}
\end{table*}

The most critical factor for executing \glspl*{LLM} is the device's memory capacity \cite{alizadeh_llm_2024}, as the model must be fully loaded into memory.
iOS enforces a memory limit beyond which applications are automatically terminated, as shown in \autoref{tab:device-selection}.
We determined these limits for all devices using the \textit{iOS Memory Budget Test}\footnote{\url{https://github.com/Split82/iOSMemoryBudgetTest}} and \textit{Xcode Instruments}\footnote{\url{https://help.apple.com/instruments}}.
As a result, certain devices in the benchmark (\autoref{tab:model-selection}) were unable to run specific models despite having a better computational performance.

Due to differences in model size and device hardware capabilities, some devices were unable to execute larger models.
The iPhone 13 (iPhone14,5), with only 4~\gls*{GB} of memory, was limited to running 1-billion-parameter models, specifically Bio~Medical~1B and Llama~3.2~1B.
The small DeepSeek R1 1.5B did not run successfully on the iPhone 13 because its higher quantization (8-bit) and growing context window led to application termination.
In contrast, the iPhone 12 Pro (iPhone13,3) was limited to models with a maximum memory usage below 3.94~\gls*{GB} (BioMedical 1B, Llama 3.2 1B, DeepSeek R1 1.5B, BioMedical 3B, Llama 3.2 3B, Phi 3 mini 3.8B).
All devices with at least 8~\gls*{GB} of memory successfully ran all selected models.

The evaluation was conducted on personal devices, some of which were equipped with protective cases.
During benchmarking, the devices were not in use, and the application remained in the foreground with the screen brightness set to 0\%, connected to a charger, and with \textit{Low Power Mode} disabled.
However, in certain scenarios, the devices either continued charging or stopped charging at a threshold due to iOS's \textit{Optimized Battery Charging}\footnote{\url{https://support.apple.com/en-us/108055}} function or temperature constraints.
While this setup does not reflect controlled lab conditions, it closely aligns with the real-world conditions emphasized in this research.

\subsection{Analysis}
\label{sec:analysis}

\paragraph{Data Processing}
All evaluation data were collected using the open-source HealthBench application and subsequently transferred to a central repository.
The data processing involved importing files, removing duplicate entries while retaining the most recent records, and restructuring the data into an analyzable format.
To ensure completeness, we verified that all questions were answered for each model and device.
Finally, we computed descriptive statistics to summarize the data and extract key features relevant to our research questions.

\paragraph{Evaluation}
\label{sec:evaluation}
To evaluate the generated answers, we utilized the AMEGA benchmark script\footnote{\url{https://github.com/DATEXIS/AMEGA-benchmark}}, modifying it to align with our testing procedure.
Specifically, existing functions were adapted to evaluate only the questions and \gls*{LLM} responses, and the \textit{reask} process was removed.

The evaluation was performed using the latest available version of OpenAI’s \textit{GPT-4o} model at the time of writing (\texttt{gpt-4o-2024-08-06}).
The evaluator model (GPT-4o) was configured with the same parameters specified by~\citeauthor{fast-2024-amega-llm}, setting the temperature to zero, omitting a system prompt, and including only the prompt (task description), rating criteria, and generator response as the user message.

Reasoning models from the DeepSeek~R1 family begin with a \textit{think} phase, in which the model generates a set of intermediate reasoning steps~\cite{wei_chain--thought_2023}.
This process, encapsulated within XML tags (\texttt{<think>}) in the raw model responses, was removed to ensure the evaluation focused solely on the final model answer.

\section{Results}
\label{sec:results}
The benchmarking process accumulated a total of 108.95 hours of computation time and generated 5,991,077 output tokens on mobile devices.
Additionally, model evaluation using OpenAI's GPT-4o involved 8,331,847 input tokens and 2,670,254 output tokens for assessing generated responses.
As noted in \autoref{sec:model-selection}, the MedFound~8B model was excluded due to a lack of consistent outputs, while other models did not fit within the memory constraints of devices with smaller memory sizes.  

\subsection{Accuracy}
\begin{table}[h]
\centering
  \begin{tabular}{lllr}
    \toprule
    Type & Size & Model & Score \\
    \midrule
    \multirow[t]{6}{*}{medical} & small & BioMedical 1B & 177.2 \\
    \cline{2-4}
     & medium & BioMedical 3B & 323.0 \\
    \cline{2-4}
     & \multirow[t]{4}{*}{large} & MedLlama 8B & 109.6 \\
     &  & BioMedical 8B & 332.9 \\
     &  & Med42 8B & 490.0 \\
     &  & Aloe 8B & 490.9 \\
    \bottomrule
    \multirow[t]{8}{*}{\cellWithForcedBreakLeft{\\non-\\medical}} & \multirow[t]{2}{*}{small} & DeepSeek R1 1.5B & 179.8 \\
     &  & Llama 3.2 1B & 256.5 \\
    \cline{2-4}
     & \multirow[t]{2}{*}{medium} & Llama 3.2 3B & 400.9 \\
     &  & Phi 3 mini 3.8B & 464.6 \\
     \cline{2-4}
     & \multirow[t]{4}{*}{large} & DeepSeek R1 7B & 296.6 \\
     &  & DeepSeek R1 8B & 399.1 \\
     &  & Qwen 2 7B & 457.7 \\
     &  & Llama 3.1 8B & 464.8 \\
     \bottomrule
    \end{tabular}
    \caption{Mean AMEGA score for each model across devices.}
    \label{tab:amega-score}
\end{table}

\autoref{tab:amega-score} presents the mean AMEGA scores for each model, categorized by type and size.
The mean score was computed across all model runs on compatible devices.

The lowest-performing model, MedLlama3-v20, a medical model with 8 billion parameters, achieved an average score of 109.6 points across devices.
In contrast, the highest-scoring model, Aloe 8B (large), another medical model based on Llama 3.1, attained 490.9 points.
The second-best performer, Med42, also a medical model based on Llama 3, closely followed with an average score of 490.0 points.

Phi 3 Mini, a non-medical model with 3.8 billion parameters (medium-sized), ranked fourth with a score of 464.6 points.
Notably, this model can run with only 6~\gls*{GB} of memory, demonstrating strong performance relative to its size.

The three Bio-Medical models, spanning different parameter sizes, exhibited relatively low AMEGA scores compared to similarly sized models.
Even the largest Bio-Medical model (8B) scored only 13 points higher than its medium-sized counterpart, reaching 332.9 points, indicating minimal performance gains despite the increased parameter count.

In contrast, the baseline Llama 3.1 model, without additional medical fine-tuning, achieved a significantly higher score of 464.8 points.
This suggests that the medical fine-tuning applied to the Bio-Medical models may have negatively impacted their performance in the cases and question-focused AMEGA benchmark, resulting in a 131.9 point reduction compared to the untuned Llama 3.1 model.

All DeepSeek models with reasoning capabilities underperformed compared to their direct competitors.
The smaller DeepSeek R1 1.5B model scored 76.7 points lower than its competitor, the Llama 3.2 1B model.
Similar trends appeared among larger models, with the 7B Qwen2 DeepSeek model scoring 102.5 points below the 8B Llama variant.
However, the original Qwen2 7B model, which lacks reasoning capabilities, achieved a score exceeding 150 points—a pattern also observed in the Llama base model.

\subsection{Performance}

\begin{figure}[h]
\centering
\begin{subfigure}{0.45\linewidth}
    \centering
    \includegraphics[width=0.9\linewidth]{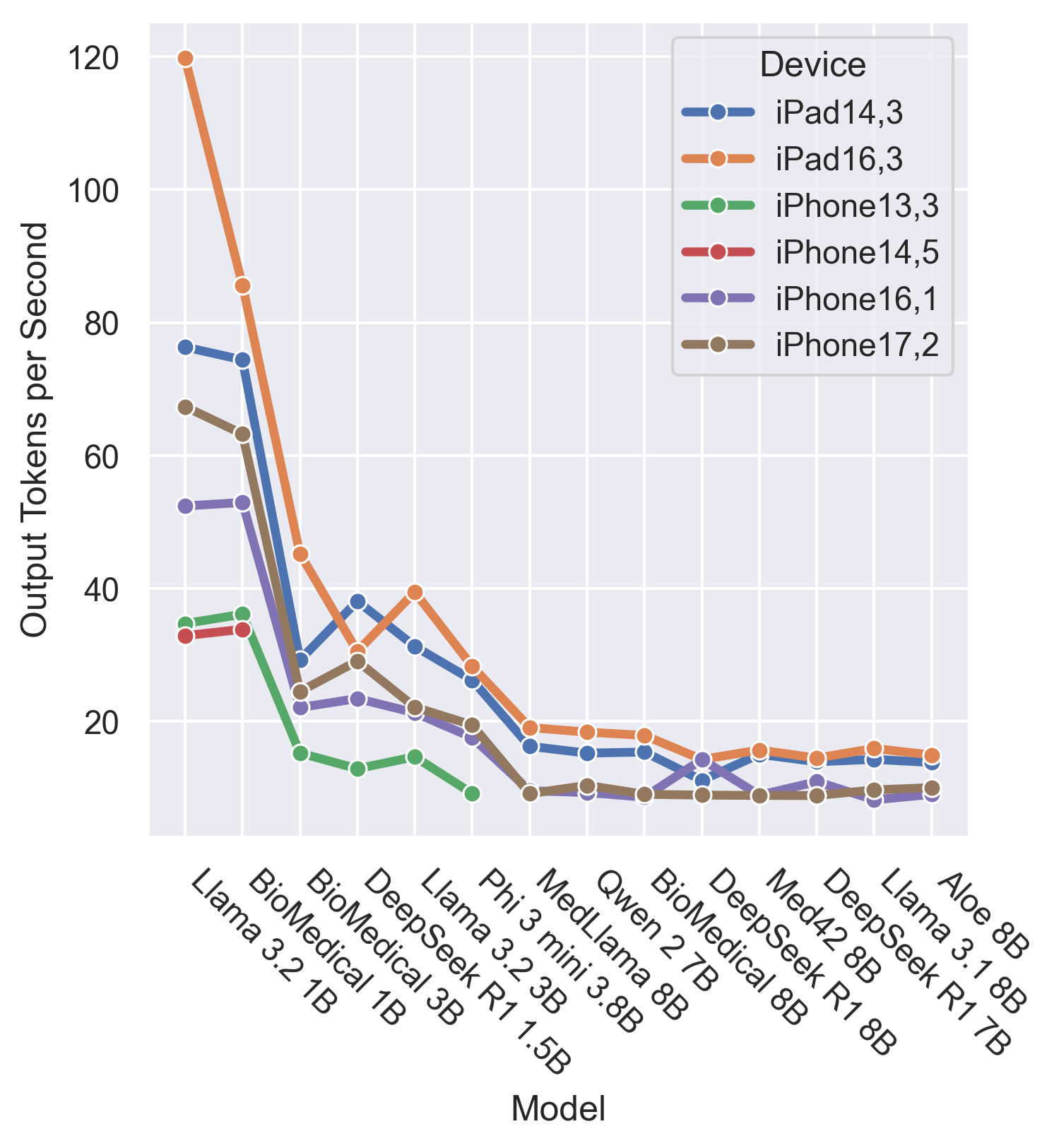}
    \caption{The output tokens per second across all devices sorted by the model's parameter size and average tokens per second.}
    \label{fig:performance:device}
\end{subfigure}
\qquad
\begin{subfigure}{0.45\linewidth}
    \centering
    \includegraphics[width=\linewidth]{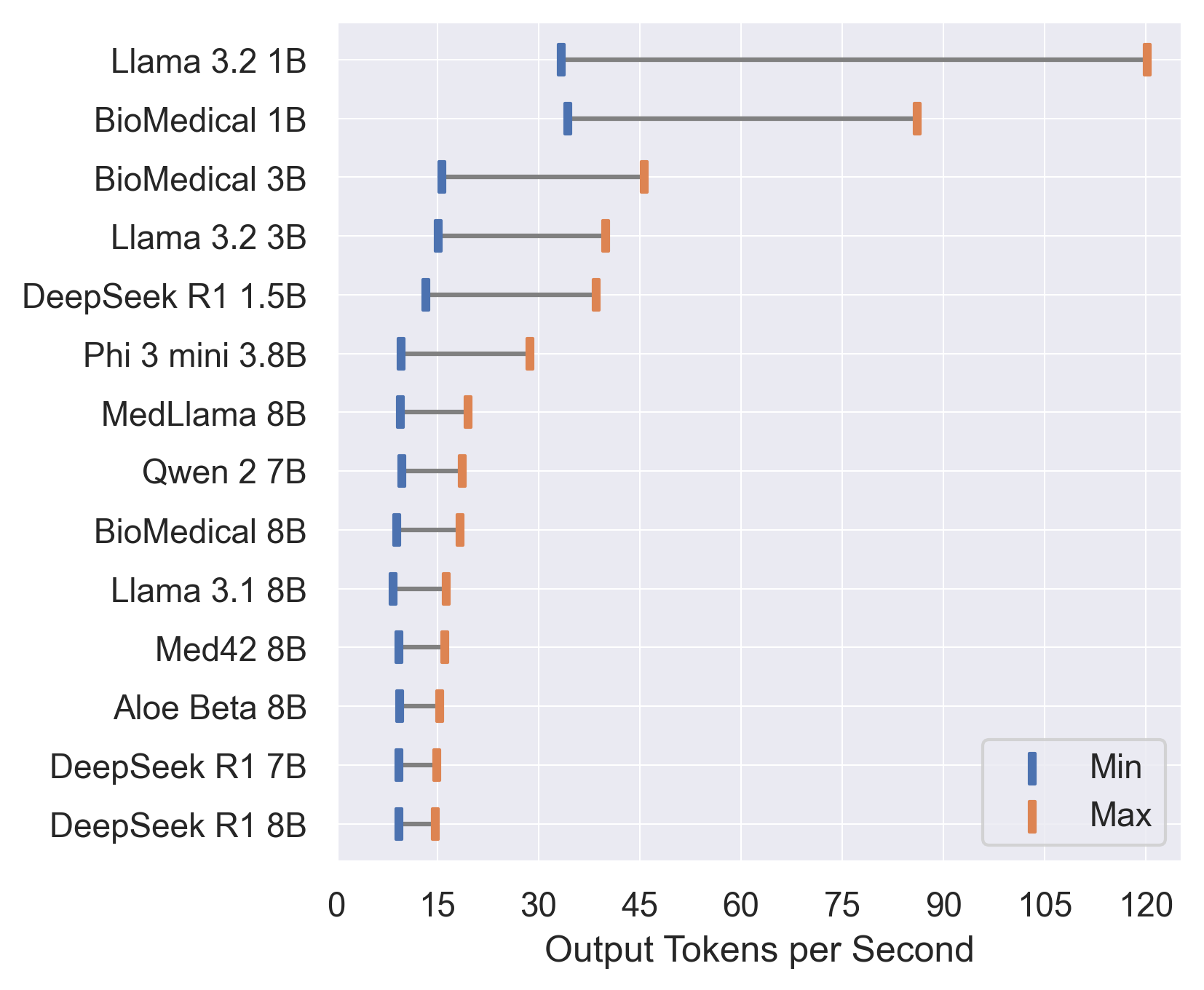}
    \vspace{5.6mm}
    \caption{The output tokens per second for each model showing the slowest and fastest performance.}
    \label{fig:performance:model}
\end{subfigure}
\caption{Performance Evaluation Across Devices and Models.}
\end{figure}

\autoref{fig:performance:device} illustrates the performance of various models across different devices.  
Among all tested devices, the iPad16,3 with 16~\gls*{GB} of memory achieved the highest throughput, reaching 120 \gls*{TPS} on the Llama 3.2 1B model.
However, its performance decreased to 30 \gls*{TPS} when running the DeepSeek R1 1.5B model.
Despite an initial performance gap among the first five models, both iPads (iPad14,3 and iPad16,3) exhibited similar performance trends across the remaining models.

The iPhone16,1 and iPhone17,2 showed performance differences on the first two models (Llama 3.2 1B and BioMedical 1B); however, their performance converged for larger models, with only minor variations.
Due to hardware limitations, the iPhone13,3 was restricted to executing smaller models, leading to missing data for models exceeding 3.8 billion parameters.

We attempted to run the DeepSeek R1 1.5B model on the iPhone14,5, as its 2.24~\gls*{GB} memory footprint suggested compatibility with the device’s 4~\gls*{GB} of memory.
However, despite meeting this requirement in theory, our investigation revealed that iOS enforces a strict memory limit, terminating applications exceeding 2.25~\gls*{GB} of usage.
This constraint ultimately prevented successful model inference.

\autoref{fig:performance:model} depicts the minimum and maximum output \gls*{TPS} for each model across all tested devices.
The minimum value reflects the average \gls*{TPS} on the slowest device, while the maximum value represents the average \gls*{TPS} on the fastest device.
Consistent with trends observed in \autoref{fig:performance:device}, models with 1B parameters generally achieved higher output speeds than larger models.

The DeepSeek R1 1.5B model, including tokens generated in the \texttt{<think>} block, performs slower than the BioMedical~3B and Llama~3.2~3B models, both of which have 3 billion parameters.  
This trend is consistent with the performance of other DeepSeek models (7B and 8B), which are the slowest overall.

The results also highlight the performance variability between the slowest and fastest devices.
A significant difference is observed between the iPad16,3 (M4, 16~\gls*{GB}, 2024) and the iPhone13,3 (A14 Bionic, 6~\gls*{GB}, 2020), particularly for smaller models.
For example, the performance gap reaches 86 \gls*{TPS} for Llama 3.2~1B.
However, this variability decreases with larger models; for the largest model both devices can execute, Phi 3 Mini 3.8B, the gap narrows to just 19 \gls*{TPS}.

\paragraph{Thermal}
\begin{figure}[htbp]
    \centering
  \includegraphics[width=0.5\linewidth]{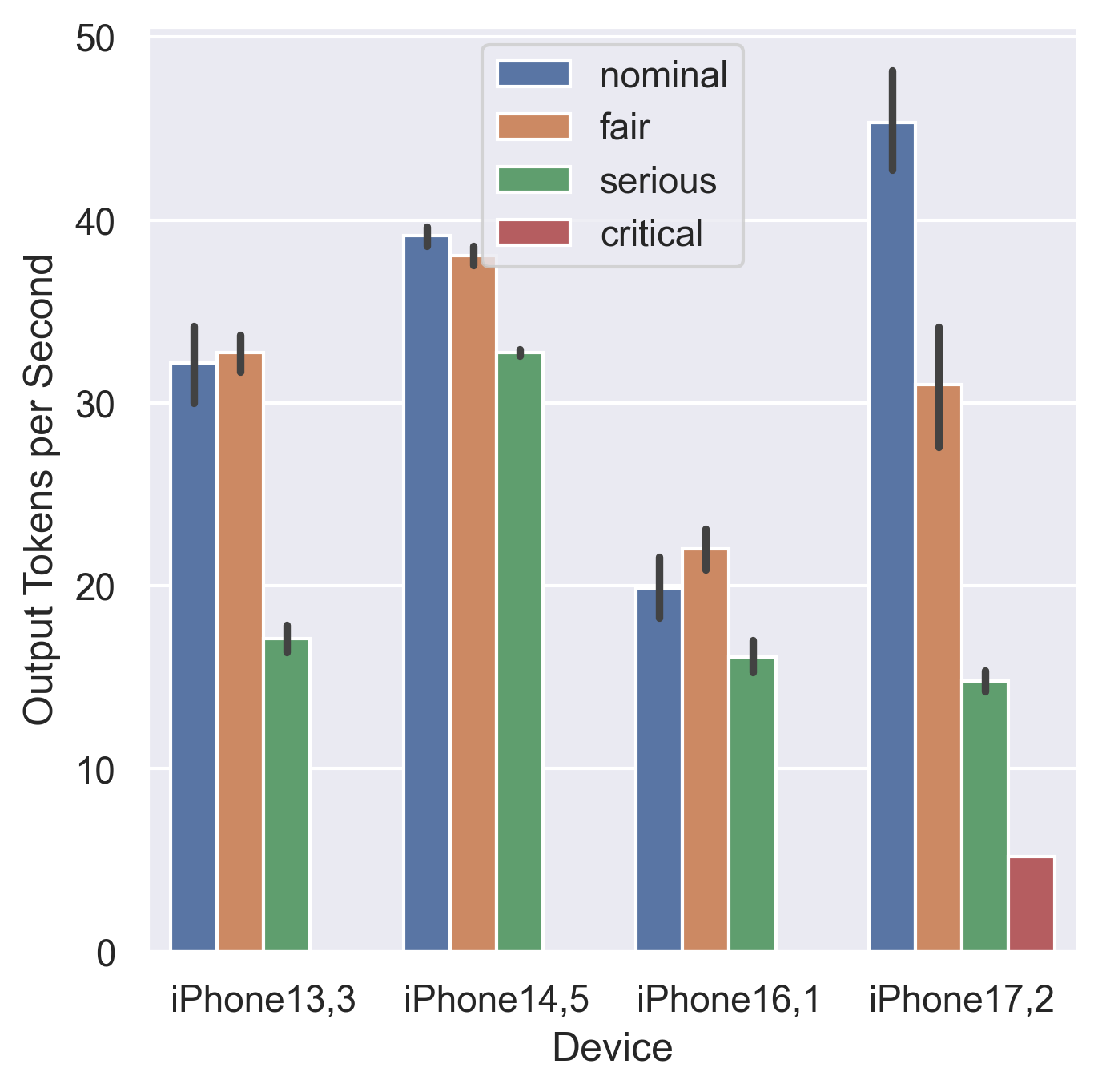}
  \caption{Output tokens per second compared to the thermal state of all iPhones over all models.}
  \label{fig:thermal}
\end{figure}

Due to the computationally intensive nature of \gls*{LLM} inference, device thermal levels were elevated compared to typical usage.
Both iPads consistently remained in the \textit{nominal} or \textit{fair} thermal state.
In contrast, the iPhones exhibited significant \gls*{TPS} variations as temperatures increased.

\autoref{fig:thermal} shows the average tokens per second across all models.
Performance remains similar in the nominal and fair states, except for the iPhone17,2.
However, a noticeable difference emerges in the \textit{serious} state, where performance drops to 52\% for the iPhone13,3, 82\% for the iPhone14,5, 72\% for the iPhone16,1, and 45\% for the iPhone17,2 compared to the \textit{fair} thermal state.
The iPhone17,2 shows a critical measurement of only 5.1~\gls*{TPS}; however, this is based on a single measurement.

\section{Discussion}
\label{sec:discussion}
This study examined the feasibility of executing \glspl*{LLM} on mobile devices by evaluating the performance of both medical and non-medical models spanning parameter sizes from 1B to 8B. Performance assessment was conducted using a state-of-the-art medical benchmark consisting of real-world open-ended questions. The impact of model inference on device performance and thermal characteristics was analyzed across hardware configurations with varying resource capabilities.

\subsection{Key Findings}
\paragraph{Models}
The results indicate that Med42 is the most suitable model for medical applications on mobile devices, achieving the highest benchmark score while requiring a minimum of 8~\gls*{GB} of memory. Aloe Beta demonstrated comparable performance, ranking closely behind Med42. However, for applications necessitating broader device compatibility, including deployment on older or lower-power hardware, Phi 3 Mini exhibited promising performance, delivering above-average output tokens per second alongside competitive accuracy.

In contrast, the MedLlama3 model demonstrated the poorest performance in the benchmark, exhibiting the least accuracy and efficiency across all tested devices despite its widespread adoption, with over 17,000 monthly downloads on Hugging Face. This suboptimal performance may be attributed to the low-temperature setting (0.1) used to ensure reproducibility, as well as potential overfitting to medical exam questions, which could limit its generalization capabilities.

The findings suggest that model size is not the sole determinant of performance. Accuracy and performance are influenced by various factors, including the quality and diversity of training data or the underlying base model architecture. However, fine-tuned models did not show noticeable performance difference compared to the general-purpose models.

Notably, the base model plays a critical role in determining on-device MLX-based inference speed. The results indicate that Llama3-based models consistently outperform Qwen2-based models, potentially due to architectural differences or varying levels of optimization in the MLX format and its Swift implementation.

The evaluation revealed recurring patterns in model outputs, with certain responses exhibiting characteristics indicative of memorized training data. In particular, some outputs resembled multiple-choice question formats, often presenting lettered answers or directly mirroring multiple-choice formats. This phenomenon is likely attributable to extensive exposure to datasets containing such question-answer formats during training.

The findings on the DeepSeek models show that all models are slower in terms of output tokens per second compared to their direct competitor, with the two largest DeepSeek models being the slowest overall. One possible explanation for this could be their larger context window, which is needed for the reasoning ('think') process.
For the accuracy, both large models score lower compared to their base models (Qwen 2 7B \& Llama 3.1 8B). A possible explanation could be that the training data include less medical and clinical information.

\paragraph{Devices}
A key finding regarding device performance is that even older, lower-performance devices are capable of running modern \glspl*{LLM} at an acceptable level. Since 2020, all high-end iPhones and iPads have been equipped with a minimum of 6~\gls*{GB} of memory, with flagship models (the iPhone Pro series) since 2023 featuring at least 8~\gls*{GB}. However, memory remains the primary limiting factor for on-device \gls*{LLM} inference. As shown in \autoref{tab:device-selection}, iPads generally exhibit a higher effective memory limit than iPhones, possibly due to differences in system resource allocation, such as the absence of background services required for cellular functionality in non-cellular iPad models.

Benchmark results indicate that \gls*{CPU} performance is not the primary bottleneck for on-device inference. With output speeds ranging from eight to ten tokens per second, performance remains sufficient for practical applications. However, certain devices, despite possessing superior hardware, do not fully utilize their capabilities to achieve significantly improved results (see \autoref{fig:performance:model}). For instance, the 2024 iPad Pro, the fastest mobile device at the time of writing, does not exhibit linear performance scaling. In some cases, its token generation speed is comparable to that of the iPhone 15 Pro. This suggests that beyond a certain threshold, hardware advancements alone do not substantially increase \gls*{TPS}, and other factors such as memory bandwidth, implementation optimizations, or base model architecture may have a more pronounced impact.

\subsection{Comparison to Literature}
To contextualize the findings, a comparison was conducted with previous studies on accuracy and performance, identifying similarities, discrepancies, and advancements introduced in this study.

At the time of writing, no published literature has utilized the AMEGA benchmark due to its recent introduction. Consequently, direct comparisons are limited to the original study by~\citeauthor{fast-2024-amega-llm}. In their evaluation, the highest-scoring model, \texttt{WizardLM-2 8x22B}, achieved an initial score of 36.3. To ensure comparability with the present results, this score was multiplied by the number of cases (20), yielding a total score of 726.0.

The highest-performing model in this study, Med42 on the iPad14,3, achieved a score of 501.9, positioning it between MedLlama-2 7B and Llama-2 7B as reported by~\citeauthor{fast-2024-amega-llm}. Among models of comparable size (8B parameters), the highest reported score was attained by Mistral 7B (634.0). The observed differences in accuracy may be attributed to several factors, including quantization, conversion to the MLX format, or variations in sampling parameters such as temperature.

In terms of performance, prior studies suggest that \gls*{TPS} on edge devices is generally lower than the values observed in this study. \citeauthor{dhar_empirical_2024} evaluated model inference on Raspberry Pi devices, reporting speeds of approximately 0.1 \gls*{TPS}. Another study~\cite{coplu_performance_2023} analyzing different iPhone models recorded speeds of up to 8 \gls*{TPS}. However, both studies employed different runtimes than MLX, which may account for the observed discrepancies.

\subsection{Limitations}
While this study offers valuable insights into the accuracy and performance of on-device \glspl*{LLM} for medical-related questions, several major and minor limitations must be acknowledged.

A primary limitation of this study is the testing setup. Four of the six devices used for evaluation were privately owned, containing actual user data and running various background applications and services. Additionally, environmental conditions such as room temperature, airflow, protective cases, charging speed, and battery level were not controlled. These factors may introduce variability in performance measurements and complicate cross-device comparisons. While a controlled benchmarking environment would enhance comparability across scientific studies, it may not accurately represent real-world on-device \glspl*{LLM} usage. This trade-off was intentionally accepted to better assess the practical applicability of the models.

Another significant limitation is the conversion of model files to the MLX format. During this process, models were quantized to 4-bit to reduce memory consumption, allowing larger models to run on resource-constrained devices. Additionally, adaptations were made, and configuration files were sanitized to ensure compatibility with the MLX framework. However, this conversion process likely led to a reduction in model performance, potentially influencing the results.

Several compromises were introduced to facilitate on-device \gls*{LLM} inference. 
According to~\cite{fast-2024-amega-llm}, the AMEGA benchmark should be conducted with a temperature of zero and without a system prompt. However, initial pilot runs using a temperature of zero resulted in degenerate outputs, prompting an adjustment to 0.1 to ensure meaningful responses. These parameter modifications can significantly influence output quality, potentially affecting both accuracy and inference speed. Additionally, similar to~\cite{fast-2024-amega-llm}, model performance evaluation was conducted using GPT-4o. A human evaluation could have identified additional issues or resulted in different answer ratings.

Finally, the model selection process was conducted without a pre-registered systematic review protocol with explicit inclusion criteria and independent validation. This may have introduced selection bias, despite efforts to include both established and emerging medical \glspl*{LLM}.

\subsection{Outlook}
\begin{figure*}[h]
    \begin{multicols}{2}
        \includegraphics[width=\linewidth]{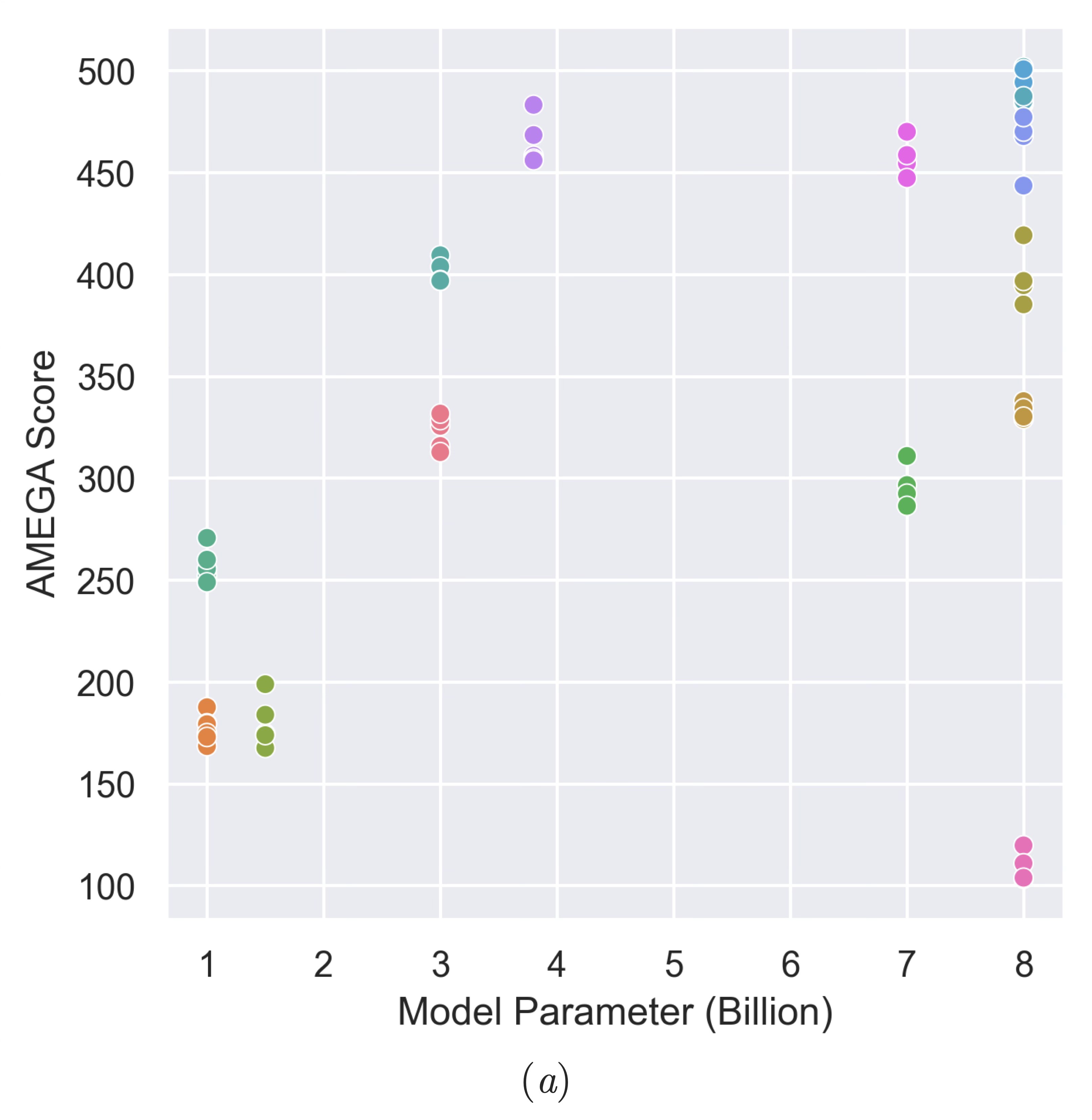}\par
        \includegraphics[width=\linewidth]{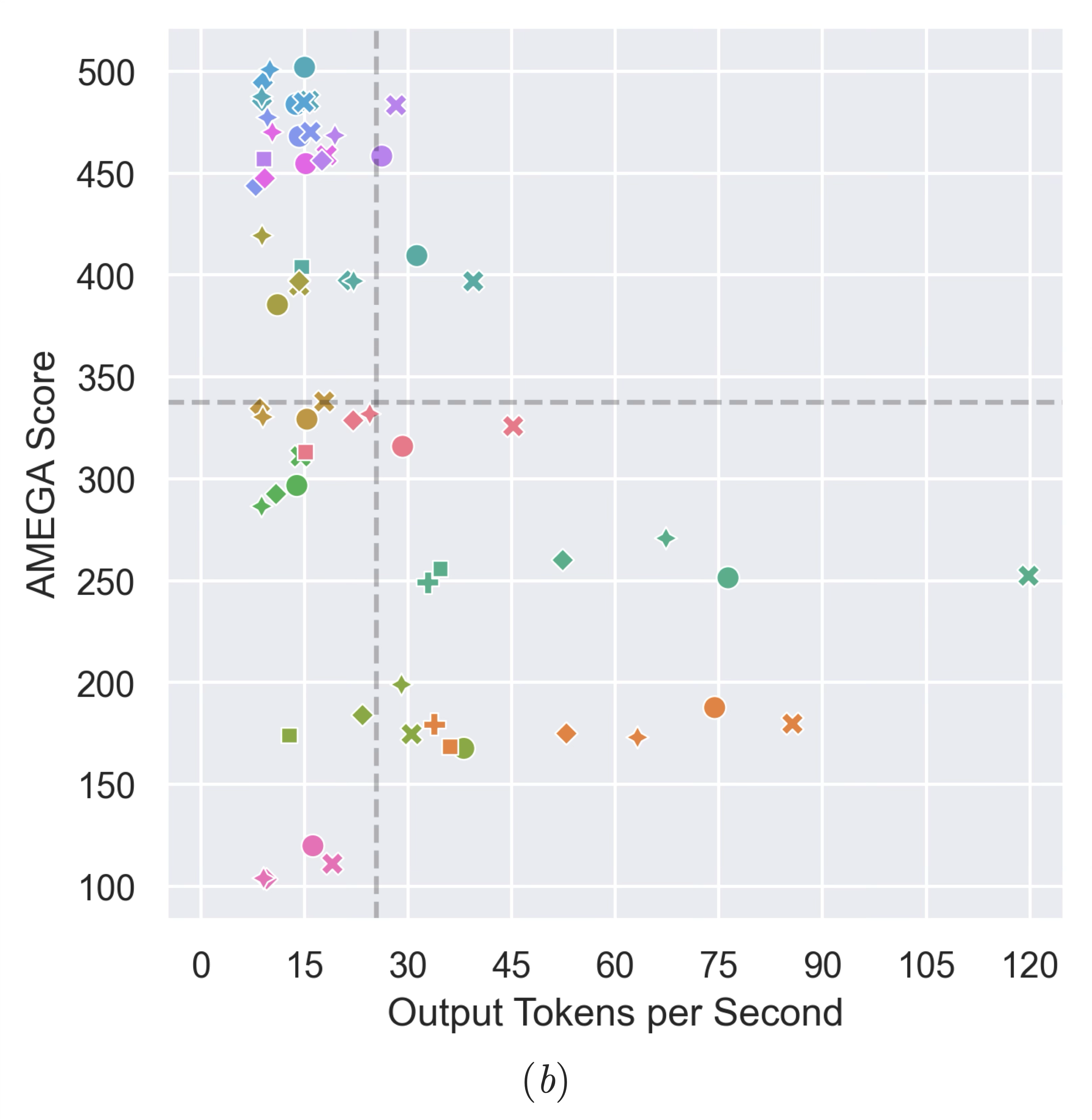}\par 
    \end{multicols}
    \begin{center}
        \includegraphics[width=0.5\paperwidth]{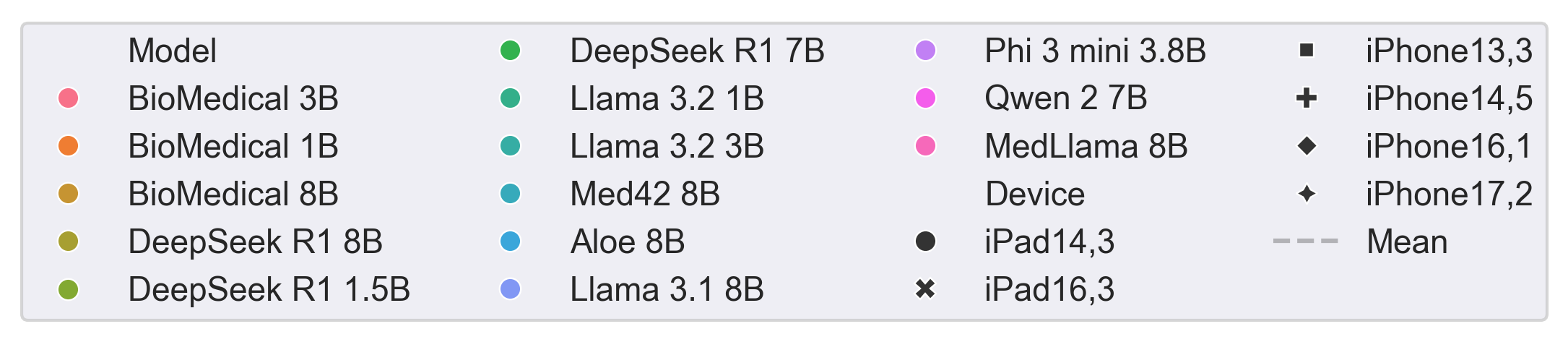}
    \end{center}
\caption{Performance comparison of \glspl*{LLM} across different devices. Plot (\textit{a}) shows the AMEGA score relative to the model's parameter size (in billions), while plot (\textit{b}) visualizes the trade-off between AMEGA score and output tokens per second. Different colors represent different models, and marker shapes indicate devices. The dashed lines highlight the mean values.}
\label{fig:outlook}
\end{figure*}

As generative AI continues to advance, enabling more personalized content, computational and memory requirements are also increasing. Privacy concerns, particularly in healthcare, may discourage users from relying on cloud-based solutions. Additionally, the growing energy consumption associated with generative AI highlights the need for more energy-efficient \glspl*{LLM} to mitigate environmental impact and reduce operational costs.

Simultaneously, \glspl*{LLM} are becoming more efficient (see \autoref{fig:outlook}b), more compact (see \autoref{fig:outlook}a), and faster (see \autoref{fig:performance:model}). These advancements facilitate the deployment of \glspl*{LLM} on privacy-focused mobile devices, reducing dependence on cloud computing and lowering operational costs. The adoption of on-device \gls*{LLM} inference is expected to expand, particularly in the medical domain, where data privacy and local processing are critical requirements.

Future research should further investigate on-device \glspl*{LLM} using real-world health-related user queries and local \gls*{FHIR}~\cite{bender2013fhir} resources, evaluating model responses based on user feedback. Additionally, energy consumption of on-device inference should be systematically analyzed, examining the impact of the chosen base model and size on efficiency in everyday use. Lastly, this study aims to encourage larger-scale research and more in-depth analyses of the effects of system prompts, temperature settings, and other configuration parameters on model accuracy and performance.

\section{Conclusion}
\label{sec:conclusion}
Our findings demonstrate that \glspl*{LLM} can operate effectively on mobile devices, offering valuable insights into health-related questions while maintaining user privacy. Both specialized medical models and general foundation models produced informative outputs with a high degree of medical accuracy. Among the evaluated models, Aloe and Med42 achieved the highest accuracy, while Phi-3 Mini exhibited a strong balance between accuracy, compact size, and inference speed.

Model performance is primarily constrained by available memory, though this limitation is expected to diminish as future devices incorporate larger memory capacities. Additionally, architectural and implementation differences may contribute to performance convergence over time. Notably, with processing speeds ranging from 11 to 30 tokens per second—exceeding the average human reading speed~\cite{brysbaert2019read}—these models exhibit practical viability for real-time mobile applications while maintaining accuracy in medical contexts.

\section*{Conflicts of Interest} 
The authors declare no conflict of interest.

\section*{Author Contributions}
\noindent\textbf{Leon Nissen}: Conceptualization, Formal analysis, Methodology, Investigation, Project administration, Software, Visualization, Writing—original draft, Writing—review \& editing, Project administration \textbf{Philipp Zagar}: Software, Writing—review \& editing \textbf{Vishnu Ravi}:  Writing—review \& editing \textbf{Aydin Zahedivash}:  Writing—review \& editing \textbf{Lara Marie Reimer}: Writing—review \& editing \textbf{Stephan Jonas}: Writing—review \& editing, Supervision \textbf{Oliver Aalami}: Writing—review \& editing, Supervision \textbf{Paul Schmiedmayer}: Conceptualization, Methodology, Investigation, Writing—review \& editing, Project administration, Supervision

\section*{Data Availability}
This study utilizes the clinical scenario AMEGA benchmark introduced by \citeauthor{fast-2024-amega-llm}.
The source code for the mobile benchmarking software as well as the modified evaluation script is available as an open-source project at: \url{https://github.com/StanfordBDHG/HealthBench}.

\section*{Acknowledgments}
Paul Johannes Kraft, Rose Hoch \& the Spezi open-source development community. 


\setcitestyle{numbers}
\citestyle{nature}
\bibliographystyle{unsrtnat}
\bibliography{main}

\end{document}